\documentclass[11pt,a4paper]{article}
\usepackage[hyperref]{emnlp-ijcnlp-2019}
\usepackage{times}
\usepackage{latexsym}

\usepackage{boldline}
\usepackage{url}
\usepackage{multirow}
\usepackage{graphicx}

\newcolumntype{C}[1]{>{\centering\arraybackslash}m{#1}}

\aclfinalcopy % Uncomment this line for the final submission

\title{Delta-training: Simple Semi-Supervised Text Classification\\
using Pretrained Word Embeddings}

\author{Hwiyeol Jo\\
  Seoul National University\\
  \texttt{hwiyeolj@gmail.com} \\
  \And
  Ceyda Cinarel\\
  Seoul National University\\
  \texttt{snu.ceyda@gmail.com} \\
}
\date{}

\begin{document}
\maketitle
\begin{abstract}
    We propose a novel and simple method for semi-supervised text classification. The method stems from the hypothesis that a classifier with pretrained word embeddings always outperforms the same classifier with randomly initialized word embeddings, as empirically observed in NLP tasks. Our method first builds two sets of classifiers as a form of model ensemble, and then initializes their word embeddings differently: one using random, the other using pretrained word embeddings. We focus on different predictions between the two classifiers on unlabeled data while following the self-training framework. We also use early-stopping in meta-epoch to improve the performance of our method. Our method, Delta-training, outperforms the self-training and the co-training framework in 4 different text classification datasets, showing robustness against error accumulation.
\end{abstract}

\section{Introduction}
\subsection{Motivation}
    Classifiers using deep learning algorithms have performed well in various NLP tasks, but the performance is not always satisfactory when utilizing small data. It is necessary to collect more data for acquiring better performance. Although collecting unlabeled text data is relatively easy, labeling in and of itself requires a considerable amount of human labor. In order to incorporate unlabeled data into a task, we have to label the data in accordance to class policies of the task, but the labeling process requires not only human labor but also domain knowledge on the classes.\\
    Semi-supervised learning~\cite{li2003learning,zhu2006semi,chapelle2009semi} can be considered a potential solution that utilizes both labeled data and unlabeled data when building a classifier. The simplest form of semi-supervised learning is self-training~\cite{yarowsky1995unsupervised}, which first builds a classifier using labeled data, and then label the unlabeled data. After which the most confident label prediction is added to training set and the process is repeated. The unlabeled data can help address data sparsity, but classification errors might be accumulated along the process.\\
    We combine self-training with the hypothesis that a classifier with pretrained word embeddings ($m_{emb}$) is always better than a classifier with randomly initialized word embeddings ($m_{rand})$, as empirically observed in various NLP tasks~\cite{turian2010word}. Our method follows the self-training framework but rather focuses on the different predictions of two sets of classifiers on unlabeled data. Therefore we can filter out incorrectly predicted data and correctly predicted data by both classifiers, which are less informative to the classifiers.
    On the other hand, differently predicted data are much more informative since much of the performance gap between the classifiers come from the different predictions. Although the differently predicted data may introduce some noise like correctly predicted by $m_{rand}$ but incorrectly predicted by $m_{emb}$, we believe that the noise is relatively small when compared with benefits.

\subsection{Contributions}
    Our contributions in this paper can be summarized as follows:
    \begin{itemize}
        \item We propose a variation of self-training framework: Delta($\Delta$)-training, which harnesses differently predicted labels between two sets of classifiers.
        \item Along with early-stopping in iterative training process, our framework outperforms the conventional self-training and co-training framework.
    \end{itemize}
    
    \begin{figure*}[t] \centering
    \includegraphics[scale=0.40]{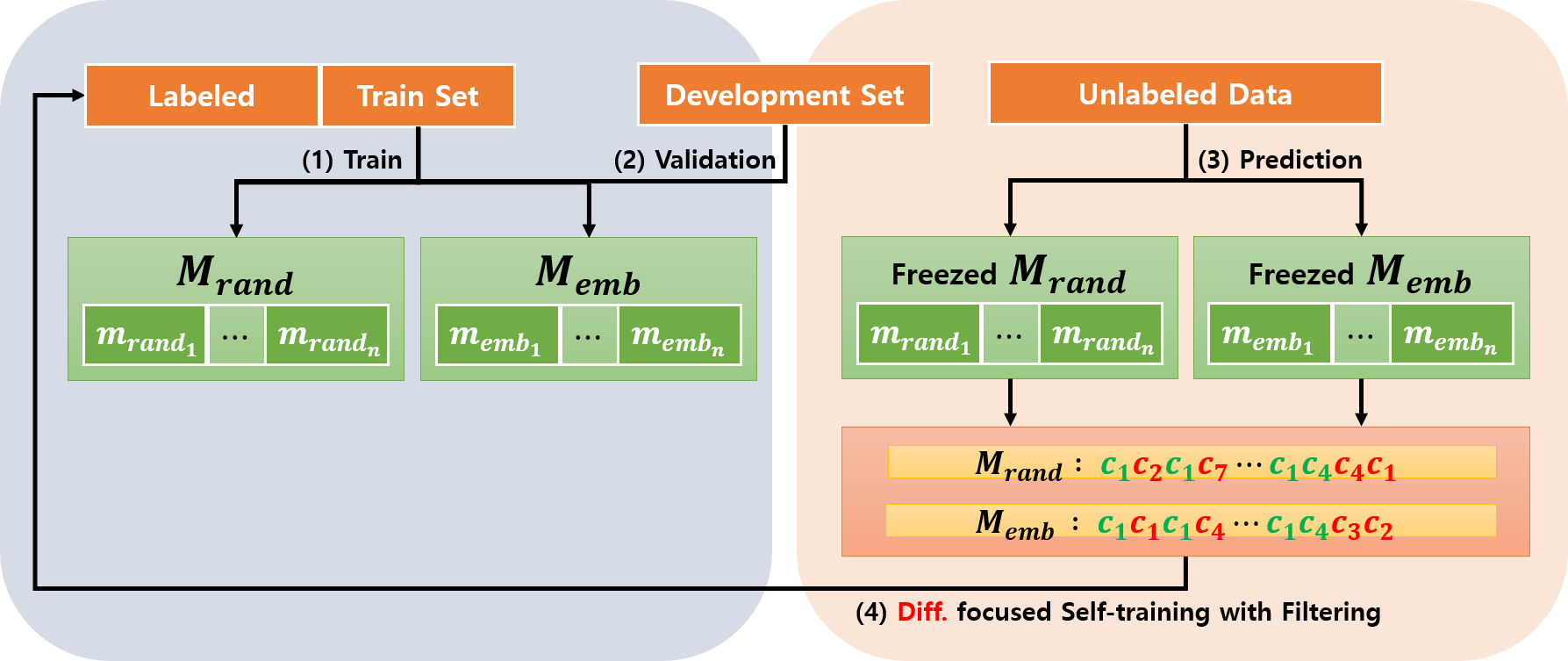}
    \caption{The flow of $\Delta$-training framework. $M_{rand}$ and $M_{emb}$ are ensembled classifiers using randomly initialized word embeddings and pretrained word embeddings, respectively. (1) We first train the sets of classifiers using training set, (2) do early-stopping using development set, (3) predict the labels of unlabeled data using the sets of classifiers trained at (1), and (4) select the high confidence labels differently predicted by each set of classifiers, adding them to training set. While following the framework, we do early-stopping in meta-epoch with the development set.}
    \label{fig:1}
    \end{figure*}
    
\section{Preliminary}
\textbf{Self-training. }
    Given labeled data $\{( x_1 , y_1 ) , \cdots , ( x_n , y_n )\}$ and unlabeled data $\{( x_{n+1} ) , \cdots , ( x_{n+l} )\}$, self-training~\cite{yarowsky1995unsupervised} first builds a model $m$ using labeled data. Next, it simply predicts the unlabeled data using pretrained model $m$. If the confidence score of the predicted label is higher than a predefined threshold $T$, then adds the label-by-prediction data to the training set. This simple approach has generated variations such as calibration~\cite{guo2017calibration}, and online learning~\cite{abney2007semisupervised}.\\ \\
\noindent\textbf{Pretrained Word Embeddings. }
    Pretrained word embeddings are based on the distributed representation hypothesis that a word can be represented as an n-dimensional vector~\cite{mikolov2013distributed}. Most of the algorithms are based on the basic idea of CBoW
    % (Continuous Bag-of-Words)
    and skip-gram. Both algorithms learn word vectors by maximizing the probability of occurrence of a center word given neighbor words or neighbor words given a center word. With this unsupervised approach, we can represent semantic and relational information.
    % which can be captured from the word order.
    The pretrained word vectors from very large corpus are used to initialize word vectors for classifiers, performing better than randomly initialized word vectors~\cite{turian2010word}.\\
    % \textbf{GloVe}~\cite{pennington2014glove} has lots of variations in respect to word dimension, number of tokens, and train sources. We used {\small\tt glove.6B} trained on Wikipedia+Gigawords and {\small\tt glove.42B.300d} trained on Common Crawl. The other pretrained GloVe do not fit in our experiment because they have different word dimension or are case-sensitive.
    %, which is developed from Google, also train the word vectors using the co-occurrence.
    % The difference is that the dot products of word vectors equal the cosine similarity between two words whereas GloVe equals the logarithm of the words' probability of co-occurrence.
    % We also use 300-dimensional \textbf{Word2Vec}~\cite{mikolov2013efficient} with negative sampling trained on GoogleNews corpus. \textbf{Fasttext}~\cite{bojanowski2016enriching} is an extension of Word2Vec, which utilizes subword information to represent an original word. We used 300-dimensional pretrained Fasttext trained on Wikipedia ({\small\tt wiki.en.vec}), using skip-gram.

\noindent\textbf{Model Ensemble. }
    Model ensemble~\cite{opitz1999popular} is using a combination of models to increase accuracy and get confidence scores on predictions. There are two types of ensemble methods, bagging and boosting. Bagging averages the predictions over a collection of classifiers whereas boosting weights the vote with a collection of classifiers.

\section{Proposed Method: $\Delta$-training}\label{sec:3}
    The overall process of our framework is illustrated in Figure~\ref{fig:1}.
    
\subsection{Different Prediction focused Self-training}
    Our method consists of two classifiers: one is randomly initialized ($m_{rand}$; random network), and the other is using pretrained word vectors ($m_{emb}$; embedded network). When ensembling, we duplicate the same classifier, $M_{rand} = ( m_{rand_1}, \cdots , m_{rand_n})$ and $M_{emb} = (m_{emb_1}, \cdots, m_{emb_n})$, respectively.\\
    We adopt bagging to increase the cases that (1) both $m_{rand}$ and $m_{emb}$ predict the labels of data correctly, and (2) $m_{rand}$ predicts incorrectly but $m_{emb}$ predicts correctly. Model ensemble is also used to pick out label-by-prediction data with high confidence, which will be used for self-training. Intuitively, the benefits of $\Delta$-training are maximized when the performance gap between two sets of classifiers is large. Also, we can ensure the performance gap not only by using pretrained embeddings but also through the ensemble setting.\\
    First, we train the classifiers using the training set with early-stopping, and return their predictions on unlabeled data. We consider the predictions of $M_{emb}$ on the unlabeled data as label-by-prediction since $M_{emb}$ always outperforms $M_{rand}$ according to our hypothesis. The hypothesis will be confirmed in Section~\ref{sec:7}.\\
    After labeling the unlabeled data, we select the data with conditions that (a) each ensembled classifiers are predicting the same class, and (b) the predictions of $M_{rand}$ and $M_{emb}$ are different. Condition (a) helps to pick out the data labeled with high confidence by the classifiers and Condition (b) helps to pick out the data which is incorrect in $M_{rand}$ but correct in $M_{emb}$. The ratio in which labels might be correct in $M_{rand}$ but incorrect in $M_{emb}$ is relatively small than vice versa (will be also presented in Section~\ref{sec:7}). We add the selected data and its pseudo-label by $M_{emb}$ to training set, and then train the classifiers again from the very first step to validate our hypothesis. If we do not start from the very first step, it might cause $M_{emb}$ to overfit and perform worse than $M_{rand}$.\\
    We denote one such iterative process, training and pseudo-labeling, as a {\em meta-epoch}.
    % One iterative process, training and pseudo-labeling, is denoted as {\em meta-epoch}.
    
% \subsection{Label Refinement}
%     While following the aforementioned training method, we need to refine the data added to the training set at the early stages. The label-by-prediction data added to training set at early stages are predicted on relatively small data than at later stages, so their labels are relatively low confidence. To deal with such problem, we perform label refinement, which returns a part of the training set labeled by prediction to label them again. We select the data to be refined sequentially and circularly. For example, after refining the data from index $0$ to $n$, we select the next data from $n+1$ to $m$. If $m$ is bigger than the size of labeled data ($N$), we select the data from $n+1$ to $N$, and return to $0$ and select the remainder.\\
%     As the meta-epoch increases, which means predicting the label with higher confidence level than the previous meta-epoch, the amount of data to refine has to be smaller. Therefore, we define the size of data to refine at every meta-epoch as {\tt\small [\#differently predicted data]/[meta-epoch]}.
\subsection{Early-Stopping in Meta-Epoch}
    Using the development set in every meta-epoch, we do early-stopping during the different prediction focused self-training. As $M_{rand}$ keeps learning based on the predictions of $M_{emb}$, the size of data which is incorrectly predicted by $M_{rand}$ but correctly predicted by $M_{emb}$ will decrease. Likewise, the size of data which is correctly predicted by both $M_{rand}$ and $M_{emb}$ will increase. Therefore, after early-stopping in meta-epoch, we simply add all the unlabeled data with its pseudo-labels to the training set. In this way, we can fully benefit from different prediction focused self-training and save training time.
    
\section{Experiment Data}
% \subsection{Pretrained Word Embeddings}
    We use \textbf{GloVe}~\cite{pennington2014glove} {\small\tt glove.42B.300d} as word embedding for $M_{emb}$. We also perform word vector post-processing method, \textbf{extrofitting}~\cite{jo2018expansional}, to improve the effect of initialization with pretrained word embeddings on text classification, as described in their paper.\\
    % We also report the performance gap in Appendix~\ref{appendix:a}.\\
    % \subsection{Text Classification Datasets}
    We use 4 text classification datasets; {\bf IMDB review}~\cite{maas2011learning}, {\bf AGNews}, {\bf Yelp review}~\cite{zhang2015character}, and {\bf Yahoo!Answers}~\cite{chang2008importance}. 

    \begin{figure*}[ht] \centering
    \includegraphics[scale=0.25]{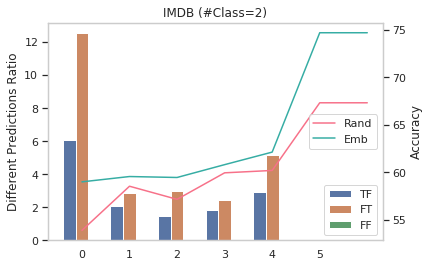}
    \includegraphics[scale=0.25]{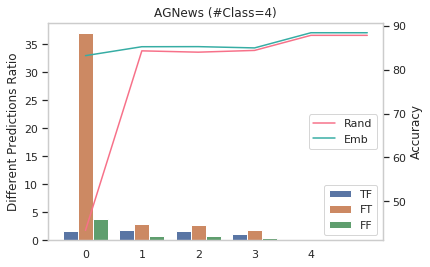}
    \includegraphics[scale=0.25]{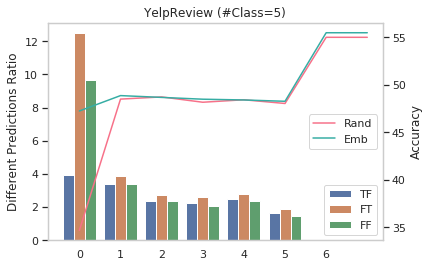}
    \includegraphics[scale=0.25]{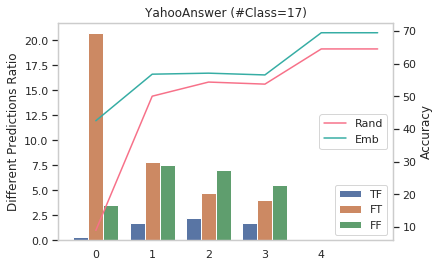}
    \caption{The training curve and the ratio of differently predicted label during $\Delta$-training. The x-axis indicates a training process (meta-epoch). TF, FT, and FF denote that correctly predicted by $M_{rand}$ but incorrectly predicted by $M_{emb}$, incorrectly predicted by $M_{rand}$ but correctly predicted by $M_{emb}$, and incorrectly predicted by both $M_{rand}$ and $M_{emb}$, respectively. We extend the end of performance lines to indicate final accuracy.}
    \label{fig:2}
    \end{figure*}
    
    \begin{figure*}[ht] \centering
    \includegraphics[scale=0.28]{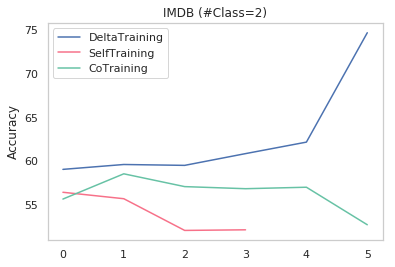}
    \includegraphics[scale=0.28]{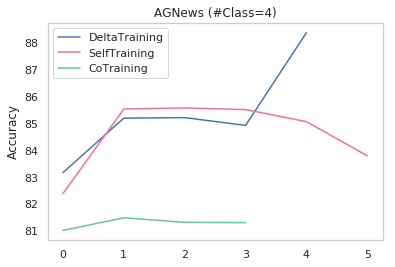}
    \includegraphics[scale=0.28]{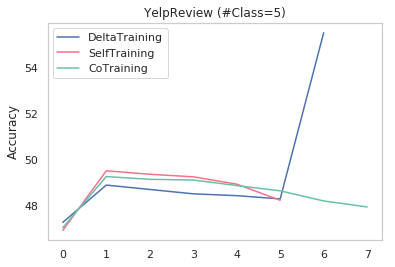}
    \includegraphics[scale=0.28]{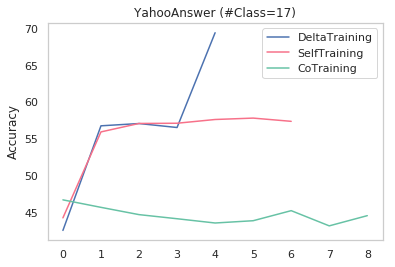}
    \caption{The performance of $\Delta$-training compared with self-training and co-training. Our method largely improves the performance in binary classification but slightly show degraded performance prior to early-stopping in meta-epoch. Then, $\Delta$-training finally brings significant performance gain after meta-level early stopping, training on all the remaining pseudo-labeled data.}
    \label{fig:3}
    \end{figure*}
    
    \begin{figure*}[ht!] \centering
    \includegraphics[scale=0.28]{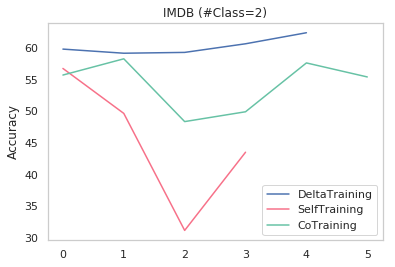}
    \includegraphics[scale=0.28]{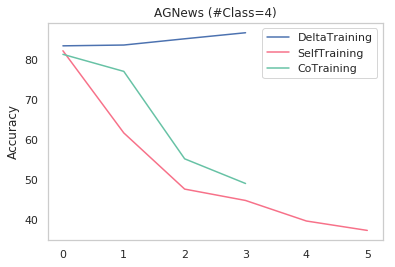}
    \includegraphics[scale=0.28]{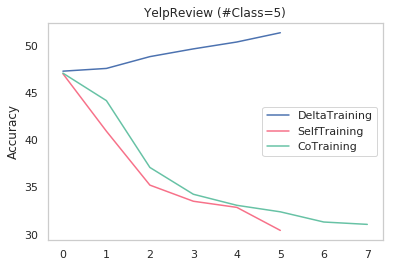}
    \includegraphics[scale=0.28]{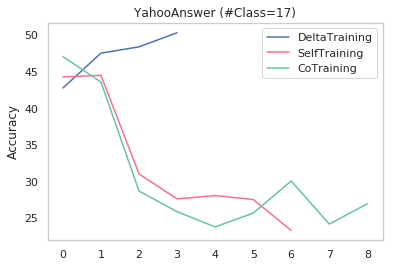}
    \caption{The performance of $\Delta$-training and other frameworks on unlabeled data. We recover the removed true labels and track the model performance. $\Delta$-training is robust against error accumulation.}
    \label{fig:4}
    \end{figure*}
    
    % \begin{figure*}[ht!] \centering
    % \includegraphics[scale=0.25]{plot_DeltaTraining_Performance_IMDB.png}
    % \includegraphics[scale=0.25]{plot_DeltaTraining_Performance_IMDB.png}
    % \includegraphics[scale=0.25]{plot_DeltaTraining_Performance_IMDB.png}
    % \includegraphics[scale=0.25]{plot_DeltaTraining_Performance_IMDB.png}
    % \caption{The effect of model ensemble?}
    % \label{fig:4}
    % \end{figure*}
    
\section{Experiment}
\subsection{Data Preparation}\label{sec:5.1}
    To emulate the environment with only a few labeled training examples, we take only 1\% of the original training set and remove the label of the remaining training set, which will be referred to as unlabeled data. Next, we assign 15\% of the training set to the development set. The development set is used to determine early-stopping for every epoch and meta early-stopping for every meta-epoch. The split data size is presented in Table~\ref{tab:1}. IMDB reviews dataset has own unlabeled data but we do not use them in order to track and report on the difference between predicted labels and true labels.
    
    \begin{table}[t] \centering
    \begin{tabular}{|c||C{1.1cm}|C{1.1cm}|C{1.1cm}|C{1.1cm}|}
    \hlineB{3}
            & \begin{tabular}{@{}c@{}} {\small\bf IMDB} \\ {\small\bf Review} \end{tabular} & {\small\bf AGNews} & \begin{tabular}{@{}c@{}} {\small\bf Yelp} \\ {\small\bf Review} \end{tabular} & \begin{tabular}{@{}c@{}} {\small\bf Yahoo} \\ {\small\bf Answer} \end{tabular} \\
            \hlineB{3}
        {\small Train} & \small 212 & \small 1,020 & \small 5,525 & \small 1,136 \\
        \hline
        {\small Test} & \small 25,000 & \small 7,600 & \small 50,000 & \small 23,595 \\
        \hline
        {\small Dev} & \small 38 & \small 180 & \small 975 & \small 201 \\
        \hline
        {\small Unlabed} & \small 24,750 & \small 118,800 & \small 643,500 & \small 132,366 \\
        \hlineB{3}
        {\small \#Class} & \small 2 & \small 4 & \small 5 & \small 17 \\
        \hlineB{3}
        \end{tabular}
        \caption{The data split information of text classification datasets. We use 1\% of the training data and remove the labels of the remaining training data, using them as unlabeled data.}
        \label{tab:1}
    \end{table}
    
\subsection{Classifier}
    We select TextCNN~\cite{kim2014convolutional} as our classifier. Due to the simple but high performance nature of TextCNN, the model can represent deep learning classifiers, and is easy to ensemble as well. We use the first 100 words of data in 300 dimensional embedding space. The model consists of 2 convolutional layers with the 32 channels and 16 channels, respectively. We adopt multiple sizes of kernels--2, 3, 4, and 5, followed by ReLU activation~\cite{hahnloser2000digital} and max-pooled. We concatenate them after every max-pooling layer. We train the model using Adam optimizer~\cite{kingma2014adam} with 1e-3 learning rate.
    % The ablation study with respect to the number of model ensemble is presented in Appendix~\ref{appendix:c}.
\subsection{Ensemble Settings}
    In the experiment, the number of embedded model ensemble ($M_{emb}$) is 3 and we do not ensemble random model ($M_{rand} = m_{rand}$) for simplicity. The baselines also use the same ensemble settings for fair comparison. 

\section{Related Works}
    $\Delta$-training is closely related to {\bf Self-training}~\cite{yarowsky1995unsupervised}, and {\bf Tri-training with Disagreement}~\cite{sogaard2010simple}. Tri-training~\cite{zhu2006semi} uses 3 classifiers to vote their classification results and labels them if all the classifiers agree with the prediction. Its extension, Tri-training with Disagreement, also uses 3 classifiers but the method utilizes a disagreement that pseudo-labels on unlabeled data if two classifiers agree with the labels but one classifier disagrees with the labels. The differences with our method respectively are (1) we harness different predictions of classifiers, and (2) we use a single model architecture where word embeddings are initialized differently.\\
    The existing semi-supervised solutions using 2 classifiers such as {\bf Co-training}~\cite{blum1998combining} cannot be fully compared with ours for (2) that a single architecture should be used. The method is built on 2 different classifiers as having different views on data, and harnesses one's pseudo-labels to train the other classifier. Instead, we imitate the co-training as if $M_{rand}$ and $M_{emb}$ have different views on the data.\\
    Refer to \citeauthor{ruder2018strong}'s work (\citeyear{ruder2018strong}) for further knowledge on those related works.
    
    \begin{figure*}[h!] \centering
    \includegraphics[scale=0.28]{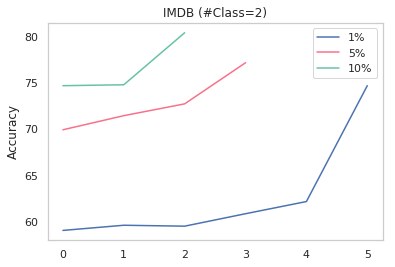}
    \includegraphics[scale=0.28]{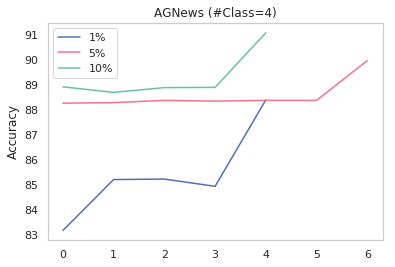}
    \includegraphics[scale=0.28]{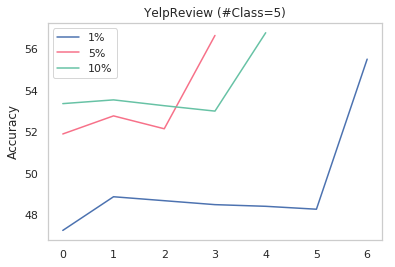}
    \includegraphics[scale=0.28]{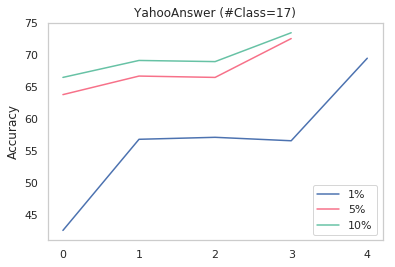}
    \caption{The performance of $\Delta$-training with respect to initial training data size. $\Delta$-training performs well in different training data size and is more useful when the training data is scarce.}
    \label{fig:5}
    \end{figure*}
    
\section{Result}\label{sec:7}
    The training curve and the ratio of differently predicted labels are presented in Figure~\ref{fig:2}. The training curves at x=0, at which point $\Delta$-training is not applied yet, confirms our hypothesis--a classifier with pretrained word embeddings always outperforms the same classifier with randomly initialized word embeddings. Also, the ratio in which labels are correct in $M_{rand}$ but incorrect in $M_{emb}$ is relatively small (TF$<$FT) than vice versa. The ratio in which labels are incorrect both in $M_{rand}$ and $M_{emb}$ (FF) changes according to baseline accuracy and the number of classes.\\
    The performance of $\Delta$-training compared with self-training and co-training is presented in Figure~\ref{fig:3}. Our method largely outperforms the conventional self-training and co-training framework in binary class classification. In multi-class classification, picking different predictions is less effective because the data could be incorrectly predicted by both $M_{emb}$ and $M_{rand}$. Therefore, after the early-stopping in meta-epoch, we simply add all the unlabeled data with its pseudo-labels to training set, which finally brings significant performance gain. In Figure~\ref{fig:4}, we observe that the performance of self-training and co-training decreases in unlabeled data after a few meta-epochs because of accumulated classification errors. On the other hand, our method is robust against error accumulation. As a result, the process of adding all the unlabeled data with its pseudo-labels to training set starts from enhanced and robust models.\\
    % so it makes the early-stopping in meta-epoch effective.\\
    % Besides, during $\Delta$-training processes, the two models guide and enhance each other so the following labeling process starts with the enhanced models.\\
    We also report the effect of initial training data size in Figure~\ref{fig:5}. The result shows that $\Delta$-training is more useful when the training data is scarce and also $\Delta$-training works well even when there is relatively more data.

% \section{Ablation Studies}\label{sec:7}
% \subsection{The Effect of Label Refinement}
%     We use YahooAnswer dataset for time complexity.
% \subsection{The Effect of the size of Initial Training set}
%     see Appendix~\ref{appendix:a}
% \subsection{The Effect of Model Ensemble}
%     see Appendix~\ref{appendix:b}

% \subsection{Score-based Semi-Supervised learning}
\section{Conclusion}
    In this paper, we propose a novel and simple approach for semi-supervised text classification. The method follows the conventional self-training framework, but focusing on different predictions between two sets of classifiers. Further, along with early-stopping in training processes and simply adding all the unlabeled data with its pseudo-labels to training set, we can largely improve the model performance. Our framework, $\Delta$-training, outperforms the conventional self-training and co-training framework in text classification tasks, showing robust performance against error accumulation.

\section*{Acknowledgments}
    The authors would like to thank Sang-Woo Lee, Woo-Young Kang, Dae-Soo Kim, and Byoung-Tak Zhang for helpful comments. Also, we greatly appreciate the reviewers for critical comments.\\
    This work was partly supported by the Korea government (2015-0-00310-SW.StarLab, 2017-0-01772-VTT, 2018-0-00622-RMI, 2019-0-01367-BabyMind, 10060086-RISF, P0006720-GENKO), and the ICT at Seoul National University.

\bibliographystyle{acl_natbib}
\bibliography{emnlp-ijcnlp-2019}

\begin{thebibliography}{20}
\expandafter\ifx\csname natexlab\endcsname\relax\def\natexlab#1{#1}\fi

\bibitem[{Abney(2007)}]{abney2007semisupervised}
Steven Abney. 2007.
\newblock \emph{Semisupervised learning for computational linguistics}.
\newblock Chapman and Hall/CRC.

\bibitem[{Blum and Mitchell(1998)}]{blum1998combining}
Avrim Blum and Tom Mitchell. 1998.
\newblock Combining labeled and unlabeled data with co-training.
\newblock In \emph{Proceedings of the eleventh annual conference on
  Computational learning theory}, pages 92--100. ACM.

\bibitem[{Chang et~al.(2008)Chang, Ratinov, Roth, and
  Srikumar}]{chang2008importance}
Ming-Wei Chang, Lev-Arie Ratinov, Dan Roth, and Vivek Srikumar. 2008.
\newblock Importance of semantic representation: Dataless classification.
\newblock In \emph{AAAI}, volume~2, pages 830--835.

\bibitem[{Chapelle et~al.(2009)Chapelle, Scholkopf, and
  Zien}]{chapelle2009semi}
Olivier Chapelle, Bernhard Scholkopf, and Alexander Zien. 2009.
\newblock Semi-supervised learning (chapelle, o. et al., eds.; 2006)[book
  reviews].
\newblock \emph{IEEE Transactions on Neural Networks}, 20(3):542--542.

\bibitem[{Guo et~al.(2017)Guo, Pleiss, Sun, and
  Weinberger}]{guo2017calibration}
Chuan Guo, Geoff Pleiss, Yu~Sun, and Kilian~Q Weinberger. 2017.
\newblock On calibration of modern neural networks.
\newblock In \emph{International Conference on Machine Learning}, pages
  1321--1330.

\bibitem[{Hahnloser et~al.(2000)Hahnloser, Sarpeshkar, Mahowald, Douglas, and
  Seung}]{hahnloser2000digital}
Richard~HR Hahnloser, Rahul Sarpeshkar, Misha~A Mahowald, Rodney~J Douglas, and
  H~Sebastian Seung. 2000.
\newblock Digital selection and analogue amplification coexist in a
  cortex-inspired silicon circuit.
\newblock \emph{Nature}, 405(6789):947.

\bibitem[{Jo(2018)}]{jo2018expansional}
Hwiyeol Jo. 2018.
\newblock Expansional retrofitting for word vector enrichment.
\newblock \emph{arXiv preprint arXiv:1808.07337}.

\bibitem[{Kim(2014)}]{kim2014convolutional}
Yoon Kim. 2014.
\newblock Convolutional neural networks for sentence classification.
\newblock In \emph{Proceedings of the 2014 Conference on Empirical Methods in
  Natural Language Processing (EMNLP)}, pages 1746--1751.

\bibitem[{Kingma and Ba(2014)}]{kingma2014adam}
Diederik~P Kingma and Jimmy Ba. 2014.
\newblock Adam: A method for stochastic optimization.
\newblock \emph{arXiv preprint arXiv:1412.6980}.

\bibitem[{Li and Liu(2003)}]{li2003learning}
Xiaoli Li and Bing Liu. 2003.
\newblock Learning to classify texts using positive and unlabeled data.
\newblock In \emph{IJCAI}, volume~3, pages 587--592.

\bibitem[{Maas et~al.(2011)Maas, Daly, Pham, Huang, Ng, and
  Potts}]{maas2011learning}
Andrew~L Maas, Raymond~E Daly, Peter~T Pham, Dan Huang, Andrew~Y Ng, and
  Christopher Potts. 2011.
\newblock Learning word vectors for sentiment analysis.
\newblock In \emph{Proceedings of the 49th annual meeting of the association
  for computational linguistics: Human language technologies-volume 1}, pages
  142--150. Association for Computational Linguistics.

\bibitem[{Mikolov et~al.(2013)Mikolov, Sutskever, Chen, Corrado, and
  Dean}]{mikolov2013distributed}
Tomas Mikolov, Ilya Sutskever, Kai Chen, Greg~S Corrado, and Jeff Dean. 2013.
\newblock Distributed representations of words and phrases and their
  compositionality.
\newblock In \emph{Advances in neural information processing systems}, pages
  3111--3119.

\bibitem[{Opitz and Maclin(1999)}]{opitz1999popular}
David Opitz and Richard Maclin. 1999.
\newblock Popular ensemble methods: An empirical study.
\newblock \emph{Journal of artificial intelligence research}, 11:169--198.

\bibitem[{Pennington et~al.(2014)Pennington, Socher, and
  Manning}]{pennington2014glove}
Jeffrey Pennington, Richard Socher, and Christopher Manning. 2014.
\newblock Glove: Global vectors for word representation.
\newblock In \emph{Proceedings of the 2014 conference on empirical methods in
  natural language processing (EMNLP)}, pages 1532--1543.

\bibitem[{Ruder and Plank(2018)}]{ruder2018strong}
Sebastian Ruder and Barbara Plank. 2018.
\newblock Strong baselines for neural semi-supervised learning under domain
  shift.
\newblock \emph{arXiv preprint arXiv:1804.09530}.

\bibitem[{S{\o}gaard(2010)}]{sogaard2010simple}
Anders S{\o}gaard. 2010.
\newblock Simple semi-supervised training of part-of-speech taggers.
\newblock In \emph{Proceedings of the ACL 2010 Conference Short Papers}, pages
  205--208. Association for Computational Linguistics.

\bibitem[{Turian et~al.(2010)Turian, Ratinov, and Bengio}]{turian2010word}
Joseph Turian, Lev Ratinov, and Yoshua Bengio. 2010.
\newblock Word representations: a simple and general method for semi-supervised
  learning.
\newblock In \emph{Proceedings of the 48th annual meeting of the association
  for computational linguistics}, pages 384--394. Association for Computational
  Linguistics.

\bibitem[{Yarowsky(1995)}]{yarowsky1995unsupervised}
David Yarowsky. 1995.
\newblock Unsupervised word sense disambiguation rivaling supervised methods.
\newblock In \emph{Proceedings of the 33rd annual meeting on Association for
  Computational Linguistics}, pages 189--196. Association for Computational
  Linguistics.

\bibitem[{Zhang et~al.(2015)Zhang, Zhao, and LeCun}]{zhang2015character}
Xiang Zhang, Junbo Zhao, and Yann LeCun. 2015.
\newblock Character-level convolutional networks for text classification.
\newblock In \emph{Advances in neural information processing systems}, pages
  649--657.

\bibitem[{Zhu(2006)}]{zhu2006semi}
Xiaojin Zhu. 2006.
\newblock Semi-supervised learning literature survey.
\newblock \emph{Computer Science, University of Wisconsin-Madison}, 2(3):4.

\end{thebibliography}

\clearpage

\end{document}